%% file: main.tex
\documentclass{article}

\PassOptionsToPackage{numbers, compress}{natbib}

\usepackage[preprint]{neurips_2026}

\usepackage[utf8]{inputenc}
\usepackage[T1]{fontenc}
\usepackage{hyperref}
\usepackage{url}
\usepackage{booktabs}
\usepackage{amsfonts}
\usepackage{nicefrac}
\usepackage{microtype}
\usepackage{xcolor}
\usepackage{booktabs}
\usepackage{multirow}
\usepackage{graphicx}
\usepackage{natbib}
\input{Sections/appendix/appendix-preamble}

\title{UniClawBench: A Universal Benchmark for Proactive Agents on Real-World Tasks}

\author{%
  \bf Zhekai Chen\textsuperscript{1$\ast$}
  ~ \bf Chengqi Duan\textsuperscript{1}\thanks{Equal contribution, listed alphabetically}
  ~~ \bf Kaiyue Sun\textsuperscript{1$\ast$}
  ~~ \bf Bohao Li\textsuperscript{1}
  ~~ \bf Yuqing Wang\textsuperscript{1} \\
  ~~ \bf Manyuan Zhang\textsuperscript{2}\thanks{Corresponding Author}
  ~~ \bf Xihui Liu\textsuperscript{1$\dagger$}
  \\
  \textsuperscript{1}HKU MMLab ~~~
  \textsuperscript{2}Meituan  ~~~
}

\begin{document}

\maketitle

\begin{figure}[h]
    \vspace{-20pt}
    \centering
    \includegraphics[width=\textwidth]{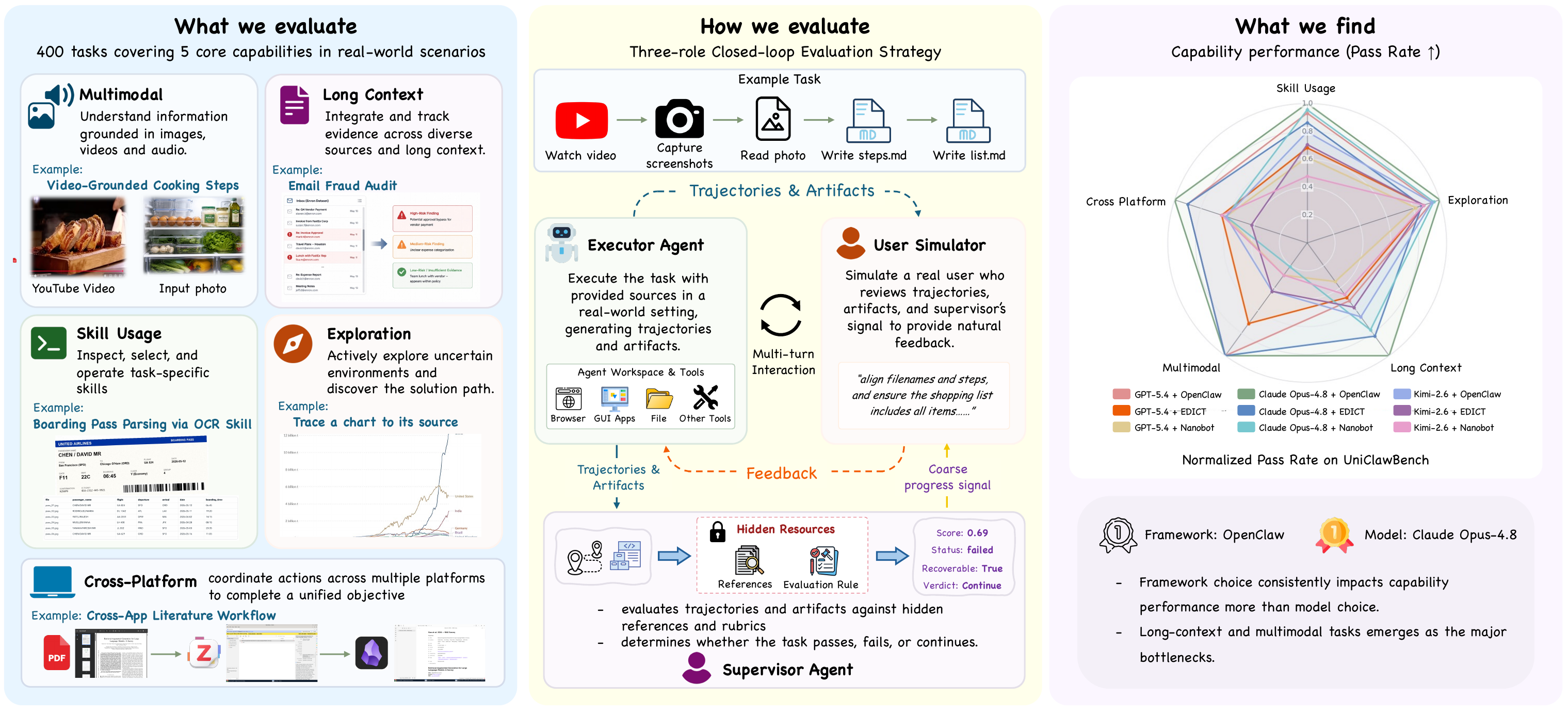}
    \caption{\textbf{Overview of UniClawBench.} UniClawBench consists of 400  bilingual real-world tasks spanning 5 core capabilities: multimodal, long-context, skill usage, exploration, and cross-platform. We propose a three-role closed-loop evaluation framework, where an executor agent performs tasks in real environments, a supervisor evaluates trajectories and artifacts using hidden rubrics, and a user simulator provides natural feedback based on executor's performance and supervisor's signal, enabling multi-turn interaction. We conduct two sets of experiments to evaluate cross-model and cross-framework performance. The capability-level results reveal that framework choice exerts a stronger influence than model choice.}
    \label{fig:teaser}
\end{figure}

\begin{abstract}
The rapid development of large language models and multimodal large language models has accelerated the emergence of proactive agents capable of operating everyday tools and assisting users in real-world environments. However, existing benchmarks struggle to evaluate such agents effectively, as they often rely on sandboxed environments and single-turn evaluation paradigms. Moreover, their scenario-based task taxonomies mix multiple model capabilities within the same task category, making it difficult to identify the root causes of agent failures. To address these limitations, we introduce UniClawBench, the first capability-driven benchmark designed to evaluate proactive agents in dynamic, real-world settings. UniClawBench is built around five foundational model capabilities: Skill Usage, Exploration, Long-Context Reasoning, Multimodal Understanding, and Cross-Platform Coordination. Based on these capabilities, we design 400 bilingual real-world tasks. Unlike previous benchmarks that rely on static, pre-recorded answers, our benchmark evaluates agents in live Docker containers using fine-grained, step-by-step completion checkpoints. Furthermore, we design a closed-loop evaluation strategy comprising an executor agent, a hidden supervisor agent, and a user agent to simulate realistic multi-turn human feedback without leaking grading criteria. To disentangle base model capabilities from framework-level design choices, we evaluate state-of-the-art models under multiple agent frameworks. Through comprehensive comparisons across both models and frameworks, we show how base model capabilities and agent framework designs jointly shape performance in real-world environments. To facilitate future research, we make our benchmark and code publicly available at \url{https://github.com/HKU-MMLab/UniClawBench}.
\end{abstract}

\input{Sections/1_introduction}

\input{Sections/2_relatedwork}

\input{Sections/3_benchmark}
\input{Sections/4_experiments}
\input{Sections/5_conclusion}

\bibliography{main}
\bibliographystyle{plainnat}

\newpage
\appendix
\input{Sections/appendix/appendix}

\end{document}

%% file: Sections/appendix/appendix-preamble.tex
\usepackage{tcolorbox}
\usepackage{makecell}
\tcbuselibrary{listings, breakable, skins}

\newtcblisting{clawcode}[1][]{%
  listing only,
  breakable,
  enhanced,
  colback=black!2,
  colframe=black!55,
  boxrule=0.4pt,
  arc=2pt,
  outer arc=2pt,
  left=6pt, right=6pt, top=4pt, bottom=4pt,
  fonttitle=\bfseries\small,
  coltitle=black,
  attach boxed title to top left={xshift=4pt, yshift=-3pt},
  boxed title style={%
    colback=white,
    colframe=black!55,
    sharp corners=northwest,
    arc=0pt,
    boxrule=0.3pt,
    top=1pt, bottom=1pt, left=4pt, right=4pt,
  },
  listing options={%
    basicstyle=\ttfamily\small,
    breaklines=true,
    breakatwhitespace=true,
    breakautoindent=false,
    breakindent=0pt,
    prebreak={},
    postbreak={},
    columns=fullflexible,
    keepspaces=true,
    upquote=true,
    showstringspaces=false,
    aboveskip=4pt, belowskip=4pt,
  },
  #1
}

\newtcblisting{clawsnippet}{%
  listing only,
  breakable,
  enhanced,
  colback=black!2,
  colframe=black!40,
  boxrule=0.3pt,
  arc=1.5pt,
  left=5pt, right=5pt, top=3pt, bottom=3pt,
  listing options={%
    basicstyle=\ttfamily\footnotesize,
    breaklines=true,
    columns=fullflexible,
    keepspaces=true,
    upquote=true,
    showstringspaces=false,
  },
}

\newtcolorbox{clawschema}[1][]{%
  breakable,
  enhanced,
  colback=black!2,
  colframe=black!55,
  boxrule=0.4pt,
  arc=2pt,
  outer arc=2pt,
  left=6pt, right=6pt, top=4pt, bottom=4pt,
  fonttitle=\bfseries\small,
  coltitle=black,
  attach boxed title to top left={xshift=4pt, yshift=-3pt},
  boxed title style={%
    colback=white,
    colframe=black!55,
    sharp corners=northwest,
    arc=0pt,
    boxrule=0.3pt,
    top=1pt, bottom=1pt, left=4pt, right=4pt,
  },
  #1
}

%% file: Sections/1_introduction.tex
\section{Introduction}
Large language models and multimodal large language models have evolved
from text-centric assistants into autonomous agents capable of executing
complex, multi-step tasks in real-world
environments~\citep{yao2022react,schick2023toolformer}.
This evolution has given rise to proactive
agents, which assist user by directly control everyday tools like browsers and terminals.
Representative platforms such as OpenClaw~\citep{openclaw}
and Nanobot~\citep{nanobot} have moved beyond isolated task execution toward
24-7 personal assistance, effectively serving as digital coworkers
that handle the full spectrum of daily work. These agents are tasked with navigating complex, real-world scenarios that span multiple modalities and platforms. This rapid evolution raises a pressing question: How can we systematically measure their true capabilities in real-world settings?

Significant efforts have been devoted to building agent benchmarks~\cite{liu2023agentbench,mialon2023gaia}. However, we identify three structural limitations that prevent them from adequately evaluating modern proactive agents. First, existing benchmarks struggle to capture real-world complexity. They often rely on self-hosted website mirrors (e.g., WebArena~\cite{zhou2023webarena}) or cached pages within virtual machines (e.g., OSWorld~\cite{xie2024osworld}), creating a profound gap between sandboxed performance and actual real-world capabilities. Second, they mostly leverage a single-turn evaluation paradigm, completely ignoring the closed-loop nature of human-agent interaction where users frequently provide iterative feedback. Finally, and most critically, existing benchmarks organize tasks by application scenario (e.g., ``office'', ``research''). This approach conflates fundamentally different abilities. When a model fails an ``office'' task, it is unclear whether the bottleneck lies in visual perception, long-context reasoning, or tool usage. Consequently, a capability-driven benchmark is urgently needed to pinpoint the precise source of failure. 

However, building a benchmark that operates in real-world environments and incorporates closed-loop user interaction introduces two fundamental challenges. The first is the lack of a stable ground truth. The expected outputs in sandboxed benchmarks are fixed when the benchmark is created, but this assumption becomes fragile in unpredictable real-world scenarios. For example, an Amazon product page may list the same product at \$199 today and \$179 tomorrow, making a pre-recorded answer quickly outdated. The second challenge lies in faithfully simulating user interaction without compromising evaluation reliability. To create a faithful closed-loop environment, an automated user simulator must dynamically provide natural corrections based on the agent's intermediate outputs. However, if this simulator is granted access to the underlying ground truth or grading criteria to guide the agent, it risks inadvertently leaking the solution or evaluation rules to the agent under test. Balancing natural, multi-turn human feedback with strict information isolation to prevent data contamination remains a significant technical challenge. 

In this paper, we introduce UniClawBench, the first capability-driven benchmark designed to evaluate proactive agents in dynamic real-world environments. UniClawBench systematically decomposes tasks along five foundational capability dimensions: Skill Usage, Exploration, Long Context, Multimodal, and Cross-Platform. To ensure comprehensive coverage, we manually design 400 bilingual tasks (English and Chinese) that encompass a wide range of authentic daily usage scenarios. All tasks run inside Docker containers equipped with real software, live browsers and local file systems. UniClawBench introduces two novel assessment mechanisms. First, to address the dynamic nature of live environments where stable ground truth is absent, we abandon pre-recorded answers. Instead, we design fine-grained, step-by-step completion checkpoints for each task. This empowers an automated supervisor with hidden evaluation criteria, ensuring it knows precisely which intermediate reasoning and execution evidence to assess at each step. Second, to solve the information-isolation paradox, UniClawBench implements a strict three-role closed-loop evaluation that acts as an information firewall. A hidden supervisor privately checks the agent's progress, while a user simulator responds only to the visible trajectory and a coarse progress signal. This setup lets UniClawBench capture realistic multi-turn interactions without revealing the underlying answers or grading criteria.

We conduct two groups of experiments to benchmark cross-model and cross-framework performance. First, we evaluate a wide range of state-of-the-art models under OpenClaw framework, which allows us to isolate differences in base model capabilities. Second, we evaluate representative models across three agent frameworks: OpenClaw~\cite{openclaw}, EDICT~\cite{edict} and Nanobot~\cite{nanobot}. This allows us to study how framework design affect performance on different tasks. Together, these comparisons reveal not only which models are more capable, but also how different agent architectures amplify or limit these capabilities in complex real-world tasks.

In summary, our main contributions are as follows:
\begin{itemize}
    \item We introduce \textbf{UniClawBench}, the first capability-driven benchmark that evaluates proactive agents in dynamic real-world environments. UniClawBench is built around five foundational capabilities, based on which we manually design 400 bilingual tasks to diagnose the root causes of agent failures.
    
    \item We propose a closed-loop evaluation strategy that better reflects real-world user-agent interaction. It introduces a three-role design with an executor agent, a hidden supervisor agent, and a user simulator agent, enabling multi-turn feedback while preventing answer or grading-criteria leakage.
    
    \item We evaluate state-of-the-art models across multiple agent frameworks. This joint analysis reveals how base model capabilities and framework designs each contribute to agent performance across different capability dimensions.

\end{itemize}

%% file: Sections/2_relatedwork.tex
\section{Related Work}

\textbf{Large language models as autonomous agents.}
Recent progress in large language models has enabled autonomous agents
that interleave reasoning, action, observation, and tool use~\citep{chen2022program,duan2026codeplot,paranjape2023art,parisi2022talm,wei2022chain,yang2023mm}.
ReAct~\citep{yao2022react} established a representative
reasoning--acting loop, while Toolformer~\citep{schick2023toolformer}
showed that language models can learn to invoke external tools.
Subsequent systems extended this paradigm toward richer orchestration and
longer-horizon problem solving: HuggingGPT~\citep{shen2023hugginggpt}
uses an LLM as a controller over specialized models, ToolLLM~\citep{qin2023toolllm}
studies large-scale API selection and invocation, and
Reflexion~\citep{shinn2023reflexion} improves agents through verbal
feedback and memory. Beyond text-only interaction, GUI and computer-use
agents such as CogAgent~\citep{hong2024cogagent} and
AppAgent~\citep{zhang2025appagent} extend the paradigm to screens and
mobile interfaces. These works establish the basic agent loop and tool-use
patterns, but they mainly target generic autonomy rather than persistent
personal assistance.

\textbf{Proactive agent systems.}
Proactive agents, distinct from reactive agents, initiate actions and pursue goals autonomously, a concept formalized by~\cite{wooldridge1995intelligent}. This foundational work underpins modern long-running assistants that leverage persistent memory, reusable skills, and local tool access. OpenClaw~\citep{openclaw} offers a robust hub-and-spoke architecture for multi-platform integration. Nanobot~\citep{nanobot} emphasizes an ultra-lightweight, extensible core for research and easy modification. Hermes Agent~\citep{hermesagent2026} features a built-in learning loop for autonomous skill creation and continuous self-improvement, building a deeper user model across sessions. These systems collectively highlight a shared design space for individualized assistance through advanced memory, skills, and tool management.

\textbf{Benchmark for AI Agents.}
Earlier benchmarks evaluate LLM agents across web, OS, mobile,
and enterprise environments~\citep{chezelles2024browsergym,drouin2024workarena,koh2024visualwebarena,liu2023agentbench,ma2024agentboard,mialon2023gaia,rawles2024androidworld,xie2024osworld,xu2024theagentcompany,yao2024tau,zhou2023webarena},
but they target isolated task execution rather than proactive personal
assistance. More recent benchmarks instead targets proactive
agent systems and falls into three groups. 
The first group measures end-to-end task completion in real environments. For instance, ClawBench \citep{zhang2026clawbench} evaluates write-heavy actions on live production websites to expose the gap between sandbox and real-world performance, while related efforts tackle harder OpenClaw tasks and desktop applications \citep{long2026liveclawbench,pinchbench2026,macagentbench2026}.
The second group studies skill use and adaptation. This is exemplified by SkillsBench \citep{li2026skillsbench}, which contrasts curated and self-generated skills to quantify their marginal effects, as well as subsequent research on continual skill learning \citep{xia2026metaclaw,ma2026skillclaw}.
The third group emphasizes trustworthiness and dynamic environments: ClawMark \citep{meng2026clawmark} models multi-day coworker scenarios with evolving backends, while other works formalize adversarial attacks and trajectory-aware evaluation \citep{wang2026assistant,ye2026claw}.
Despite this progress toward live execution, evaluating agents in real-world environments introduces critical instability, yet these recent benchmarks still rely on static ground-truth answers or single-turn evaluation. Furthermore, they organize tasks by scenario and lack analysis on agent frameworks. In contrast, UniClawBench tackles dynamic environments via hidden checkpoint rubrics and a closed-loop multi-turn evaluation strategy. Moreover, its capability-oriented taxonomy and cross-framework evaluation (OpenClaw, Nanobot, EDICT) explicitly disentangle intrinsic model competence from framework-level design choices.

%% file: Sections/3_benchmark.tex
\section{UniClawbench}
\label{sec:benchmark}
\begin{figure}[t]
    \centering
    \includegraphics[width=\textwidth]{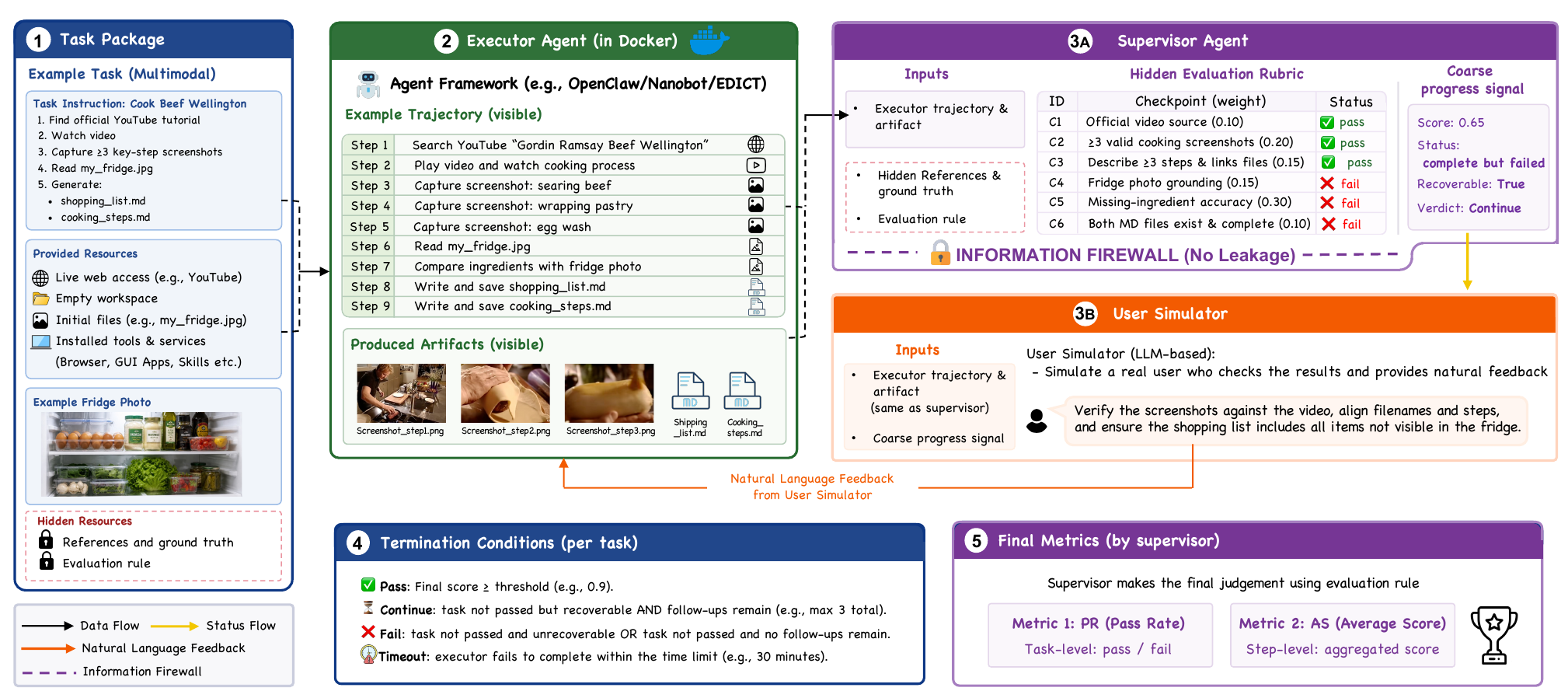}
    \caption{\textbf{Three-role Closed-loop Evaluation Strategy.} A task is executed by an agent (Executor) within a real-world environment using only public inputs (e.g., tools, files, and web access). The resulting observable trajectory and produced artifacts are assessed by a hidden Supervisor, which applies fine-grained checkpoint-based rubrics and private references to compute a structured evaluation state (e.g., pass / fail / continue) and score. To prevent leakage of evaluation criteria, only a coarse progress signal is exposed through an information firewall to a User Simulator, which generates natural-language feedback based solely on visible evidence. The sanitized feedback is returned to the executor for iterative refinement, forming a multi-turn closed loop. The example illustrates a multimodal task (Beef Wellington cooking with fridge grounding), where the agent must produce screenshots and grounded outputs,
    emphasizing strict evidence-based evaluation.}
    \label{fig:three_role_strategy}
\end{figure}

This section presents the architectural framework of UniClawBench. We first establish a capability-oriented task taxonomy and detail the benchmark composition, including task categorization, the hierarchical task-package structure, and comprehensive data statistics (\S\ref{sec:taxonomy}). Subsequently, we articulate the three-role closed-loop evaluation strategy, which decouples the evaluated executor, hidden supervisor agent, and public user simulator agent to support multi-turn feedback without leaking private grading information (\S\ref{sec:three_role}). Finally, we elaborate on the execution and evaluation system, including the docker-based runtime, task-resource injection, artifact collection, and supervisor-driven scoring pipeline (\S\ref{sec:eval-system}).

\subsection{Capability-Oriented Task Taxonomy and Benchmark Composition}
\label{sec:taxonomy}
Existing benchmarks typically organize tasks by application domain (e.g., shopping or travel) or by interaction surface (e.g., web and desktop environments). These taxonomies are useful for comparing performance across settings, but they provide limited diagnostic evidence about which capability causes an agent to fail. A failure in a domain-level task may reflect many different bottlenecks, such as incorrect tool use, insufficient exploration, weak visual grounding, poor long-context reasoning, or failure to coordinate across applications.
In realistic personal workflows, a capable proactive agent must invoke specialized tools, investigate unfamiliar or underspecified environments, maintain consistency over long evidence chains, ground its actions in non-textual signals, and coordinate state across multiple applications. These requirements motivate UniClawBench's capability-oriented taxonomy, which categorizes tasks into five core dimensions: Skill Usage, Exploration, Long-Context Reasoning, Multimodal Understanding, and Cross-Platform Coordination. Each task is assigned to the category that best captures its primary capability bottleneck, i.e., the capability whose absence would prevent successful completion.

\textbf{Skill-usage tasks.}
Skill-usage tasks evaluate an agent's proficiency in selecting, inspecting, and operating task-specific tools or APIs. The task package incorporates explicit skills, local services, fixture data, or required command-line workflows. The key bottleneck is therefore not discovering where information may exist, but correctly choosing, understanding, and executing the provided tool interface to produce a verifiable artifact. Representative tasks include OCR of boarding-pass scans, spreadsheet and CSV reconciliation, access-log analysis, audio transcription, Mermaid or diagram generation, SQLite querying, Git or Docker audits, document conversion, and API-backed workflows. These tasks are designed so that prior knowledge or a plausible final answer is insufficient: the agent must actually consult the
declared skill or tool interface, apply it to the provided data, and save structured outputs such as JSON, CSV, reports, logs, or converted files.

\textbf{Exploration tasks.}
Exploration tasks characterize an agent's capacity for open-ended investigation in the presence of incomplete, noisy, or potentially misleading information. 
Agents are required to reverse-engineer undocumented web APIs and their real fields to pull data instead of scraping HTML, audit shell or infrastructure configurations and reconcile silent conflicts, select a single best option from a fixed candidate set under quantitative constraints, trace images or data points back to their authoritative source, validate tool behavior inside local fixtures, or solve single-position puzzles and board games. Unlike skill-usage tasks, where the relevant tool interface is usually explicit, exploration tasks require the agent to determine which sources, files, services, logs, or hypotheses are
relevant before it can complete the task. These tasks emphasize the process of exploration itself: agents must
examine candidate sources, reject incorrect alternatives, document negative evidence, and leave a trace that distinguishes real investigation from a guessed conclusion. The expected outputs usually include both a final artifact and supporting audit records, such as logs, JSON notes, command transcripts, screenshots, hashes, or source comparisons.

\textbf{Long-context Reasoning tasks.}
Long-context tasks evaluate whether agents can aggregate and maintain evidence across heterogeneous sources, long documents, or extended interaction traces.  The defining difficulty is not the length of any single document, but the need to preserve constraints, reconcile conflicting evidence, and produce a globally consistent synthesis across many pieces of context. These
tasks necessitate synthesizing information from multiple web pages, videos, PDFs, reviews, emails, logs, or research materials before producing a coherent final answer. Examples include comparing headphone recommendations across Bilibili and YouTube and then checking product prices and specifications, planning subscriptions or
travel under many constraints, synthesizing reading packs into reports or slides, monitoring event streams, auditing long email or fraud records, and solving tasks that require sustained state tracking such as games or multi-step investigations. The main difficulty is not retrieving isolated facts but maintaining global consistency across scattered evidence, resolving conflicts, and
producing an auditable synthesis.

\textbf{Multimodal Understanding tasks.}
Multimodal tasks require agents to extract, interpret, and generate information grounded in images, videos, and audio. A task is assigned to this category only when the correctness of the final output depends on evidence that must be obtained from non-textual content, rather than from metadata, filenames, or
plausible textual inference alone. The tasks are constructed around realistic
content rather than synthetic visual questions. For example, agents may need to locate a specific figure in an academic paper and recreate it, reconstruct visual diagrams or SVG-like objects, inspect food or product photos, extract evidence from YouTube videos, generate subtitles or lyrics, or organize local image collections. These tasks typically require agents to combine visual perception with tool use and content generation: a successful run must not only identify the right visual content, but also save verifiable outputs such as reproduced plots, screenshots, scripts, captions, or structured summaries.

\textbf{Cross-platform Coordination tasks.}
Cross-platform tasks require agents to synchronize and orchestrate information across heterogeneous applications and interfaces. We treat cross-platform coordination as a capability rather than an environment label: the agent must preserve state, transfer information, and verify side effects across application
boundaries.
Unlike tasks that can be completed entirely in a browser or terminal, these tasks may involve moving
information among web pages, desktop GUI applications, local files, calendars,
spreadsheets, note-taking tools, citation managers, PDF readers, and
communication platforms. For example, an agent may need to read a PDF, create
a Zotero record, export a BibTeX citation, write an Obsidian literature note,
and save screenshots proving that the real desktop applications were used.
Other tasks involve travel planning across official websites and map services,
receipt reconciliation through GUI tools, calendar/event creation, or multi-
application research workflows. Success depends on preserving state and
evidence across platforms, not merely producing a textual answer.

Overall, UniClawBench contains 400 bilingual real world tasks, with 40 English and 40 Chinese tasks for each capability. All tasks are meticulously manually constructed based on genuine, day-to-day scenarios encountered by everyday users. We intentionally constrain and design each scenario so that its completion primarily bottlenecks on a single model capability, while allowing auxiliary operations to appear naturally when needed. Each task is designed as a complete task package, rather than a standalone prompt-answer pair.  A task package includes the user instruction, task-specific context, input files or web resources, tools and services, skills, expected outputs, and hidden evaluation references. A representative task example is illustrated in Figure~\ref{fig:three_role_strategy}.

\begin{figure}[t]
    \centering
    \includegraphics[width=0.9\textwidth]{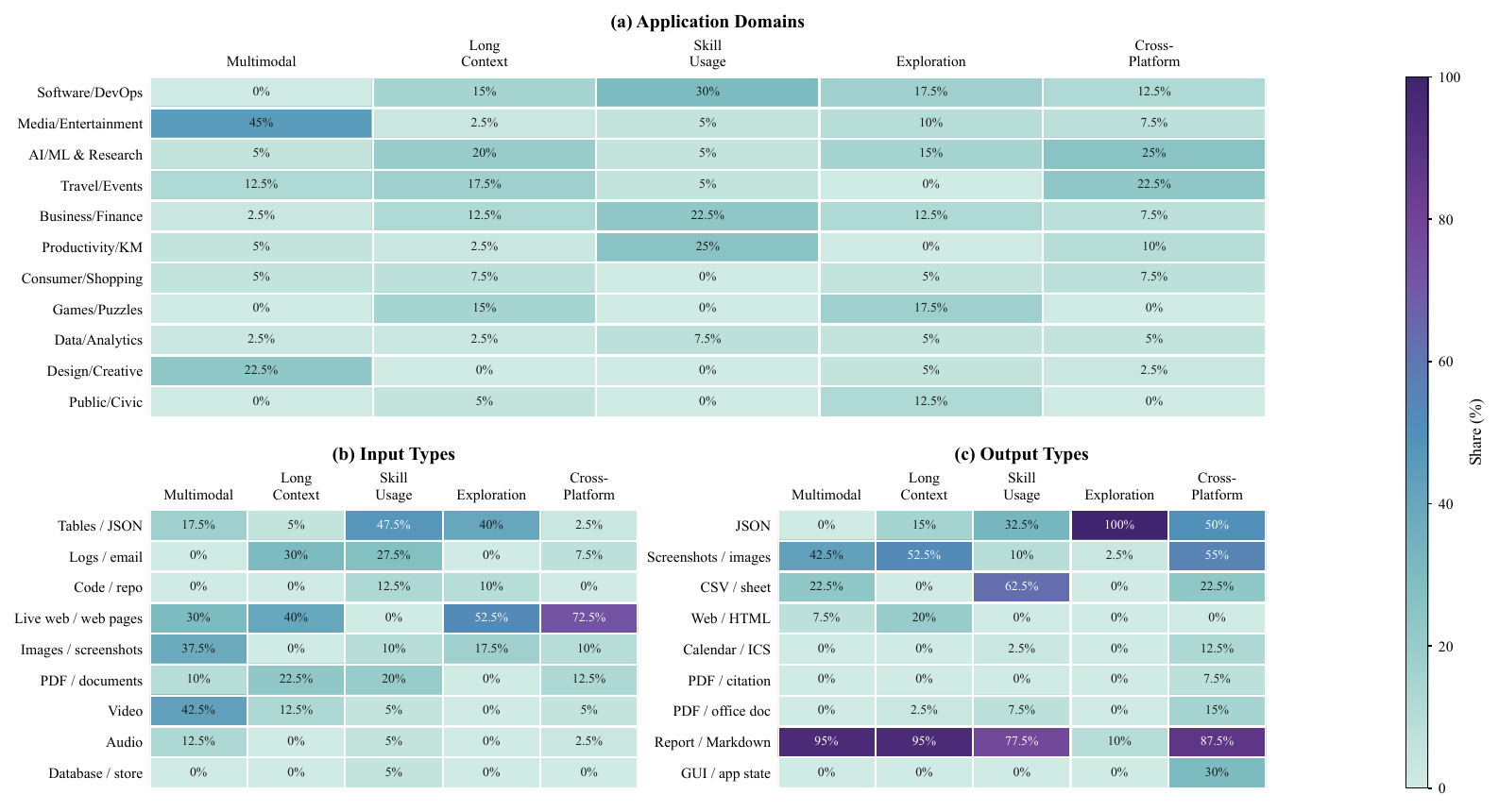}
    \caption{Diversity and composition statistics of UniClawBench. The heatmaps detail (a) the broad application-domain coverage designed to prevent scenario bias, as well as the diverse (b) input and (c) output formats. These distributions reflect the multimodal and heterogeneous nature of real-world agent tasks. Each value denotes the within-category percentage.}
    \label{fig:heatmap}
    \vspace{-15pt}
\end{figure}

To ensure a comprehensive and realistic evaluation, UniClawBench encompasses a highly diverse set of task formats and scenarios. As detailed in the Figure~\ref{fig:heatmap}, our benchmark spans a broad spectrum of application domains, ensuring that the evaluation of each foundational capability is not biased toward any specific domain. Furthermore, to faithfully reflect the heterogeneous nature of real-world tasks, the tasks incorporate a rich variety of input and output modalities. Rather than relying exclusively on standard text or JSON, agents are required to process complex inputs, such as live web pages, code repositories, videos, and databases, and successfully produce diverse, actionable outputs, including calendar files, office documents, diagrams, and GUI application states.

\vspace{-10pt}
\subsection{Three-role Closed-loop Evaluation Strategy}
\vspace{-5pt}
\label{sec:three_role}
Unlike single-turn assistants, proactive agents are typically used through multi-turn user-agent interaction. A user may inspect an intermediate result, point out missing requirements, ask for refinement, or provide additional constraints. Therefore, evaluating only the first response does not fully capture whether an agent can recover from partial progress and continue a task under user feedback. UniClawBench introduces a closed-loop evaluation design that models this interaction while preserving controlled and reproducible grading. The central challenge is to provide useful follow-up feedback without exposing the hidden evaluation criteria.

Our design involves three roles. The \textbf{executor} is the agent being evaluated. It receives only the public task instruction and the permitted task package, including available files, tools, services, and runtime resources. The \textbf{supervisor} is a hidden evaluator agent. It observes the full execution trajectory and the produced artifacts, autonomously analyzes them against task-specific hidden references and evaluation rubrics, and gives out judgments.
The \textbf{user simulator} is also an agent.
It analyzes the visible execution trajectory and outputs to generate natural feedback that the executor can act on in each round. Unlike the supervisor, the user simulator is designed to emulate the behavior of a typical end user: it focuses on visible progress, result quality, and apparent issues, rather than relying on predefined references or detailed technical inspection.

This design introduces a risk of information leakage: if the user simulator observes hidden rubrics or ground-truth answers, its feedback may reveal privileged hints to the executor. UniClawBench therefore isolates the supervisor from the user simulator. Although the supervisor has access to the exact grading criteria, it only forwards a limited high-level status signal, such as whether the attempt has passed, failed, or remains recoverable. Relying only on this status signal and the agent's visible trajectory, the user simulator generates natural feedback without access to hidden references or supervisor rationales. A feedback rewriter further sanitizes this feedback before it is delivered to the executor. This workflow
mimics realistic human-agent interaction while keeping the hidden evaluation criteria separated from the agent-facing dialogue. An illustration is shown in Figure~\ref{fig:three_role_strategy}.
\vspace{-5pt}
\subsection{Evaluation System}
\label{sec:eval-system}

UniClawBench implements the closed-loop evaluation through a host-supervised, container-executed runtime. For each task, the executor is launched in a fresh Docker environment containing the public instruction, input sources, tools, skills and relevant services. Hidden references and grading rubrics are never exposed to the executor. The executor runs inside the container using one of the supported agent systems, such as OpenClaw, Nanobot, or EDICT. It can interact with the browser, command line, GUI applications, local files, and injected services, and is instructed to save useful evidence and final outputs to a designated result directory. After each executor turn, the runner collects the observable trajectory, including transcripts, tool-use records, runtime states, and saved artifacts. These materials are then provided to the hidden supervisor together with the task-specific evaluation references. The supervisor works in a separate workspace and it scores the executor's performance using checkpoint-based rubrics and returns a structured evaluation state: pass, fail, or continue. A pass indicates that the task has been successfully completed, while a fail denotes a fundamental, irrecoverable error where the objective can no longer be achieved. A continue signifies that the agent's current progress is flawed or incomplete, but the task remains recoverable in subsequent turns. Alongside this status, the supervisor outputs a detailed justification, mapping the collected evidence to the rubrics to substantiate its decision. If the attempt is incomplete but recoverable, the public user simulator receives only the visible trajectory and the high-level status signal. It generates natural follow-up feedback, which is sanitized before being sent back to the executor as the next user message. The loop continues until the task passes, fails irrecoverably, reaches the follow-up budget, or encounters a timeout limit. This system preserves realistic multi-turn interaction while keeping grading information isolated from the evaluated agent.

%% file: Sections/4_experiments.tex
\section{Experiments}
\label{sec:experiment}
\subsection{Environmental Setup}
We conduct two groups of experiments. First, we compare ten executor models under the same OpenClaw framework~(v2026.3.11), which isolates model-level capability differences. Second, we select three representative models and evaluate them under OpenClaw (v2026.3.11), Nanobot (v0.1.5.post3), and EDICT, allowing us to study how agent-framework design affects certain performance. All compared runs use the same tasks, hidden evaluation rubrics and scoring pipeline. For both supervision and user simulation, we use separate Codex agents based on GPT-5.4 with high reasoning effort. To reduce context overhead, screenshots from the executor trajectory are not directly inserted into the initial context of the supervisor or user simulator. Instead, they are stored as observable files and can be inspected on demand when visual evidence is needed. To ensure a fair comparison and enable the system image to handle browser and GUI tasks, we abandon the OpenClaw browser extension, installing and preparing each frameworks with the same basic skills, including self-written Advanced Packaging Tool (apt) skill, DuckDuckGo Search skill~\cite{duckduckgo}, Web Search skill~\cite{websearch}, Agent-Browser skill~\cite{agentbrowser}, and Desktop Control~\cite{desktopcontrol} skill. All experiments are executed on Intel Core i7-13700 host machines with 16GB of RAM, with every isolated Docker containers allocated 2GB of memory. On average, each task run takes approximately 17.4 minutes.

In all experiments, each task allows up to two follow-up user interactions after the initial instruction. To support the agentic and interactive nature of our tasks while avoiding unbounded execution, we impose two time limits. The first is a global timeout, which bounds the total wall-clock time across all turns of a task. The second is a per-turn timeout, which limits the duration of any single executor turn. For standard tasks, we set the global timeout to 30 minutes and the per-turn timeout to 20 minutes. For long context tasks, which require processing substantially larger inputs and performing more extensive reasoning, we increase these limits to 45 minutes and 30 minutes, respectively. We report two metrics: Pass Rate (PR), the percentage of tasks that reach a final pass state, and Average Score (AS), the mean checkpoint-based score assigned by the supervisor. When a run exceeds either the global or per-turn timeout, we credit the task with the highest checkpoint-based score achieved among its completed turns, rather than discarding the partial progress made up to that point.

\vspace{-3pt}
\subsection{Benchmark Evaluation Reliability Study}

To validate the reliability of our automatic evaluation pipeline, we randomly sample 50 completed trajectories and ask three human experts to independently evaluate each of them. Each expert provides both a binary pass/fail judgment and a continuous completion score corresponding to PR and AS, respectively. For human PR, we take the majority vote among the three experts and for human AS, we average their completion scores. We then compare these aggregated human judgments with the supervisor-produced PR and AS. The automatic pass/fail decision achieves 92.0\% agreement with the human majority vote, while the checkpoint-based AS obtains a strong correlation with averaged human completion scores, with Pearson
$r=0.71$ and Spearman $\rho=0.68$.

\input{Tables/5_benchmark_model_avg}
\subsection{Cross-Model Benchmark Results}
We benchmark 10 SOTA models with Openclaw framework on the English and Chinese subset of UniClawbench. The results are presented in Table~\ref{tab:openclaw_benchmark_avg}. The results show that while leading closed-source models like Claude Opus-4.8 and GPT-5.4 achieve the highest overall pass rates, the absolute success rates remain strictly below 50\%. This highlights the extreme difficulty of the benchmark, revealing a profound gap between sandboxed capabilities and real-world task execution. Notably, open-source models are rapidly closing this performance gap. Models like Qwen-3.5-Plus and Kimi-2.6 show highly competitive results that even surpasses closed-source models like Gemini-3.1-pro. A widespread execution issue across all evaluated models is the severe ``halfway failure'' phenomenon, evidenced by the significant gap between their intermediate average scores and final pass rates. Most models consistently secure high checkpoint-based scores but ultimately fail to complete the entire task. This indicates that while current agents can successfully make partial progress, they frequently make irrecoverable errors during extended execution chains, lacking the reliability required for complex real-world tasks.  From the capability perspective, models perform relatively better on Skill Usage and Exploration tasks, where the main challenge is to identify the right tool or evidence source and apply it correctly. In contrast, Long Context, Multimodal, and Cross Platform tasks remain much harder, because they require agents to keep memories through long trajectories, ground decisions in non-textual evidence, and coordinate actions across multiple applications. This suggests that current
agents are already reasonably capable of local tool operation and information seeking, but still struggle with long-horizon memory, multimodal grounding, and cross-application coordination.

\input{Tables/6_benchmark_framework_avg}
\subsection{Cross-Framework Benchmark Results}
Table~\ref{tab:cross_framework_avg} details the performance of representative models across the OpenClaw, Nanobot, and EDICT frameworks. The results reveal that framework architecture which determines how execution trajectories are organized, profoundly influences task success. The performance gap between frameworks widens as base model capabilities increase, highlighting how structural design either amplifies or bottlenecks intrinsic model reasoning.

OpenClaw consistently achieves the highest pass rate across all tested models. As a centralized, single-agent framework, its primary advantage lies in minimal information loss. The original task constraints, intermediate tool execution evidence, and closed-loop user feedback are seamlessly preserved within a unified trajectory. This cohesive context management allows strong models to reliably translate partial progress into complete, strictly verifiable task successes.

EDICT exhibits a distinct pattern: relatively high average scores coupled with notably lower pass rates. This discrepancy exposes the ``coordination friction'' inherent in its multi-agent orchestration. The central orchestrator dispatches tasks based on discrete state polling (e.g., Kanban boards) but lacks real-time supervision over downstream sub-agents. If a sub-agent fails to transmit precise state updates or drops context during handoffs, the execution pipeline stalls, yielding partial progress that fails. Furthermore, during long-context tasks, identity and role constraints are easily forgotten. As a result, the orchestrator does the work itself instead of assigning it. The system pays the high token cost of multiple agents but gets none of the collaborative benefits.

Designed with an ultra-lightweight structure, Nanobot demonstrates a highly optimized token usage in practice, utilizing significantly fewer tokens than OpenClaw (e.g., 0.57M vs. 1.15M average input tokens for GPT-5.4). However, this extreme efficiency introduces an inherent trade-off with performance. In rigorous, real-world tasks requiring extended evidence chains and complete trajectories, this simplified context management often struggles to produce fine-grained reasoning steps and long textual evidence. Consequently, while the framework effectively achieve partial progress and secures high intermediate checkpoint scores, it frequently fails to generate the complete results demanded by strict evaluations, resulting in a notably lower overall pass rates.

\subsection{Token Usage and Performance Progression}
\begin{figure}[t]
    \centering
    \includegraphics[width=\textwidth]{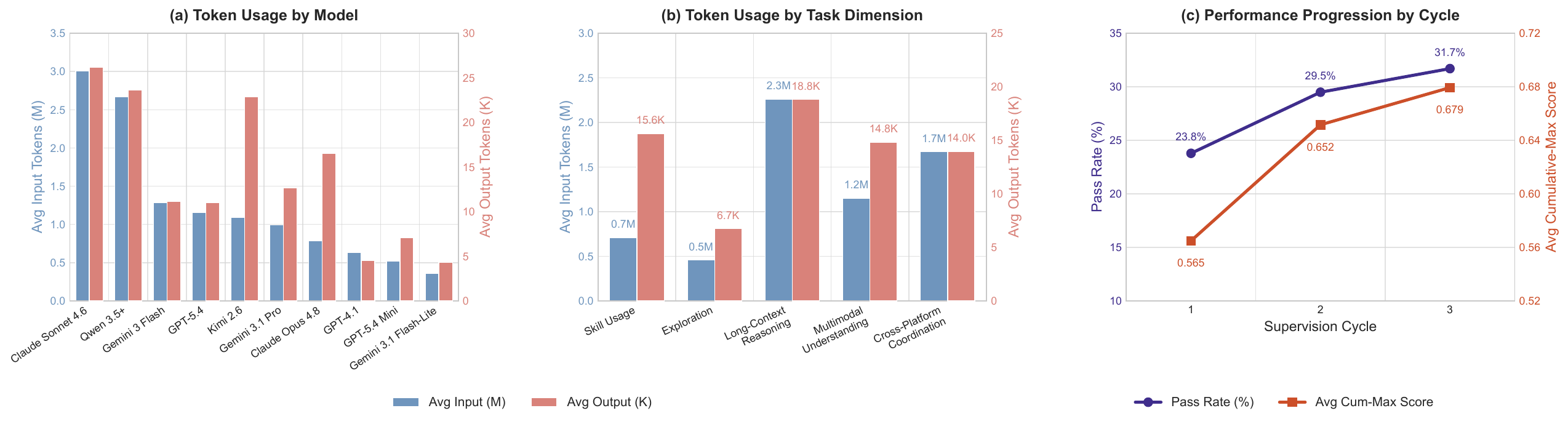}
    \caption{\textbf{Token Usage and Performance Progression by Cycle.} (a) and (b) display the token usage on the OpenClaw system, categorised by model and task dimension, (c) demonstrates the performance progression by cycle motivated by the user follow-ups.}
    \label{fig:token_and_user}
\end{figure}

As illustrated in Figure~\ref{fig:token_and_user} (a) (b), under the OpenClaw framework, token usage varies significantly across different models and inherently increases for capability dimensions where current models exhibit relative weaknesses, like long-context and multimodal understanding. Table~\ref{tab:cross_framework_avg} also provides statistics on token usage across different frameworks, where EDICT consume a huge amount of token in total and Nanobot keeps its extreme token efficiency. Furthermore, Figure~\ref{fig:token_and_user} (c) demonstrates that agent performance consistently improves across interaction cycles, validating that multi-turn user feedback is essential for dynamic error recovery. This also mimics user-agent interaction in real life.

%% file: Tables/5_benchmark_model_avg.tex
\vspace{-10pt}
\begin{table*}[htbp]
    \centering
    \caption{Average quantitative evaluation across English and Chinese subsets of UniClawBench. To isolate inherent model capabilities from framework-level design choices, all 10 models are evaluated as the executor under the same OpenClaw framework. Performance is measured by Pass Rate (PR) and checkpoint-based Average Score (AS) across the five foundational capability categories.}
    \label{tab:openclaw_benchmark_avg}
    \resizebox{\textwidth}{!}{%
    \begin{tabular}{l c c c c c c c c c c c c}
        \toprule
        \multirow{2}{*}{\textbf{Model}} & \multicolumn{2}{c}{\textbf{Skill Usage}} & \multicolumn{2}{c}{\textbf{Exploration}} & \multicolumn{2}{c}{\textbf{Long Context}} & \multicolumn{2}{c}{\textbf{Multimodal}} & \multicolumn{2}{c}{\textbf{Cross Platform}} & \multicolumn{2}{c}{\textbf{Overall}} \\
        \cmidrule(lr){2-3} \cmidrule(lr){4-5} \cmidrule(lr){6-7} \cmidrule(lr){8-9} \cmidrule(lr){10-11} \cmidrule(lr){12-13}
        & PR & AS & PR & AS & PR & AS & PR & AS & PR & AS & PR & AS \\
        \midrule
        Kimi-2.6~\cite{kimi26} & 0.438 & 0.834 & 0.775 & 0.884 & 0.287 & 0.655 & 0.075 & 0.494 & 0.237 & 0.679 & 0.362 & 0.709 \\
        Qwen-3.5-Plus~\cite{team2026qwen3} & 0.300 & 0.771 & 0.787 & 0.915 & 0.200 & 0.648 & 0.075 & 0.583 & 0.225 & 0.737 & 0.318 & 0.731 \\
        GPT-4.1~\cite{GPT-4.1} & 0.075 & 0.492 & 0.512 & 0.769 & 0.075 & 0.338 & 0.075 & 0.418 & 0.025 & 0.425 & 0.152 & 0.488 \\
        GPT-5.4-Mini~\cite{GPT-5.4-mini} & 0.188 & 0.681 & 0.688 & 0.834 & 0.100 & 0.498 & 0.062 & 0.454 & 0.037 & 0.615 & 0.215 & 0.616 \\
        Gemini-3.0-Flash~\cite{Gemini-3.0-flash} & 0.287 & 0.750 & 0.800 & 0.899 & 0.163 & 0.621 & 0.087 & 0.544 & 0.075 & 0.678 & 0.282 & 0.698 \\
        Gemini-3.1-Flash-Lite~\cite{Gemini-3.1-flash-lite} & 0.050 & 0.569 & 0.600 & 0.835 & 0.062 & 0.374 & 0.037 & 0.409 & 0.025 & 0.530 & 0.155 & 0.543 \\
        Claude Sonnet-4.6~\cite{claude-sonnet-4.6} & 0.512 & \textbf{0.873} & 0.812 & 0.913 & 0.400 & 0.649 & \textbf{0.212} & \textbf{0.626} & 0.338 & 0.753 & 0.455 & 0.763 \\
        Gemini-3.1-Pro~\cite{Gemini-3.1-pro} & 0.300 & 0.769 & \textbf{0.850} & \textbf{0.947} & 0.188 & 0.646 & 0.163 & 0.583 & 0.125 & 0.690 & 0.325 & 0.727 \\
        GPT-5.4~\cite{GPT-5.4} & 0.512 & 0.845 & 0.775 & 0.897 & 0.225 & \textbf{0.705} & 0.175 & 0.617 & 0.350 & \textbf{0.804} & 0.407 & \textbf{0.774} \\
        Claude Opus-4.8~\cite{claude-opus-4.6} & \textbf{0.550} & 0.829 & 0.825 & 0.928 & \textbf{0.438} & 0.576 & 0.175 & 0.493 & \textbf{0.388} & 0.686 & \textbf{0.475} & 0.702 \\

        \bottomrule
    \end{tabular}%
    }
    \vspace{-10pt}
\end{table*}

%% file: Tables/6_benchmark_framework_avg.tex
\begin{table*}[t]
  \centering
  \caption{Average cross-framework performance across the English and Chinese subsets of UniClawBench. We evaluate three representative models across OpenClaw, EDICT, and Nanobot frameworks. Results are presented as Pass Rate (PR), Average Score (AS), and average executor-side
  input/output token (Token$_{I}$ \& Token$_{O}$) usage per task. Input tokens are reported in millions and output tokens in thousands.}
  \label{tab:cross_framework_avg}
  \resizebox{\textwidth}{!}{
  \begin{tabular}{ll cc cc cc cc cc cccc}
  \toprule
  \multirow{2}{*}{\textbf{Model}} & \multirow{2}{*}{\textbf{Framework}} & \multicolumn{2}{c}{\textbf{Skill Usage}} & \multicolumn{2}{c}{\textbf{Exploration}} & \multicolumn{2}{c}{\textbf{Long Context}} & \multicolumn{2}{c}{\textbf{Multimodal}} &
  \multicolumn{2}{c}{\textbf{Cross Platform}} & \multicolumn{4}{c}{\textbf{Overall}} \\
  \cmidrule(lr){3-4} \cmidrule(lr){5-6} \cmidrule(lr){7-8} \cmidrule(lr){9-10} \cmidrule(lr){11-12} \cmidrule(lr){13-16}
  & & PR & AS & PR & AS & PR & AS & PR & AS & PR & AS & PR & AS & Token$_{I}$ & Token$_{O}$ \\
  \midrule

  \multirow{3}{*}{GPT-5.4~\cite{GPT-5.4}}
  & OpenClaw~\cite{openclaw} & \textbf{0.512} & \textbf{0.845} & \textbf{0.775} & \textbf{0.897} & \textbf{0.225} & \textbf{0.705} & \textbf{0.175} & \textbf{0.617} & \textbf{0.350} & \textbf{0.804} & \textbf{0.407} & \textbf{0.774} & 1.15 & 11.0 \\
  & EDICT~\cite{edict}    & 0.375 & 0.791 & 0.725 & 0.883 & 0.212 & 0.693 & 0.125 & 0.559 & 0.250 & 0.792 & 0.338 & 0.744 & 1.68 & 18.3 \\
  & Nanobot~\cite{nanobot}  & 0.338 & 0.684 & 0.675 & 0.819 & 0.150 & 0.569 & 0.050 & 0.499 & 0.237 & 0.630 & 0.290 & 0.640 & 0.57 & 9.5 \\

  \midrule

  \multirow{3}{*}{Claude Opus-4.8~\cite{claude-opus-4.6}}
  & OpenClaw~\cite{openclaw} & \textbf{0.550} & \textbf{0.829} & \textbf{0.825} & \textbf{0.928} & \textbf{0.438} & 0.576 & \textbf{0.175} & 0.493 & \textbf{0.388} & 0.686 & \textbf{0.475} & \textbf{0.702} & 0.78 & 16.4 \\
  & EDICT~\cite{edict}    & 0.475 & 0.751 & 0.713 & 0.838 & 0.362 & \textbf{0.583} & \textbf{0.175} & \textbf{0.576} & 0.350 & \textbf{0.687} & 0.415 & 0.687 & 2.15 & 42.7 \\
  & Nanobot~\cite{nanobot}  & 0.525 & 0.766 & 0.787 & 0.867 & 0.338 & 0.548 & 0.037 & 0.264 & 0.237 & 0.488 & 0.385 & 0.587 & 0.49 & 16.2 \\

  \midrule

  \multirow{3}{*}{Kimi-2.6~\cite{kimi26}}
  & OpenClaw~\cite{openclaw} & \textbf{0.438} & \textbf{0.834} & \textbf{0.775} & \textbf{0.884} & \textbf{0.287} & \textbf{0.655} & \textbf{0.075} & 0.494 & \textbf{0.237} & 0.679 & \textbf{0.362} & \textbf{0.709} & 1.09 & 22.9 \\
  & EDICT~\cite{edict}    & 0.388 & 0.813 & 0.725 & 0.868 & 0.250 & 0.601 & \textbf{0.075} & \textbf{0.506} & 0.163 & \textbf{0.686} & 0.320 & 0.695 & 2.53 & 51.5 \\
  & Nanobot~\cite{nanobot}  & 0.263 & 0.684 & 0.725 & 0.847 & 0.200 & 0.501 & 0.050 & 0.317 & 0.150 & 0.519 & 0.278 & 0.573 & 0.86 & 21.3 \\

  \bottomrule
  \end{tabular} }
  \end{table*}
\vspace{-8pt}

%% file: Sections/5_conclusion.tex
\section{Conclusion and Limitations}
\label{sec:conclusion}
In this paper, we introduce UniClawBench, a capability-driven benchmark designed to evaluate proactive agents in dynamic, real-world environments. Moving beyond static sandboxes, we decompose agent evaluation into five foundational capability categories across 400 bilingual real-world tasks. To faithfully capture multi-turn human-agent collaboration without leaking grading criteria, we propose a novel three-role closed-loop framework comprising an executor agent, a hidden supervisor agent, and a user simulator agent. Our evaluations reveal that current agents struggle with long-horizon memory and cross-platform coordination, and that framework architectures often bottleneck intrinsic model capabilities. While UniClawBench introduces a rigorous evaluation framework, it is limited by the relatively small set of 400 manually curated tasks, possible instability from live-environment execution, and the potential bias that may be introduced by LLM-based evaluation. Despite these constraints, UniClawBench successfully pinpoints the root causes of agent failures, paving the way for more reliable proactive assistants.

%% file: Sections/appendix/appendix.tex
):\MessageBreak
    \space\space\string\input{Sections/appendix/appendix-preamble}%
  }{%
    The preamble file loads tcolorbox and defines the boxed-listing\MessageBreak
    environments used by sections A--D. See appendix/README.md.%
  }{}%